\documentclass{article}

\usepackage[preprint, nonatbib]{neurips_2026}

\usepackage[utf8]{inputenc}
\usepackage[T1]{fontenc}
\usepackage{hyperref}
\usepackage{url}
\usepackage{booktabs}
\usepackage{amsfonts}
\usepackage{amsmath}
\usepackage{amssymb}
\usepackage{amsthm}
\usepackage{nicefrac}
\usepackage{microtype}
\usepackage{xcolor}
\usepackage{graphicx}

\newtheorem{theorem}{Theorem}
\newtheorem{lemma}{Lemma}
\newtheorem{corollary}{Corollary}
\newtheorem{proposition}{Proposition}
\theoremstyle{definition}
\newtheorem{definition}{Definition}

\newcommand{\E}{\mathbb{E}}
\newcommand{\R}{\mathbb{R}}
\newcommand{\Prob}{\mathbb{P}}
\newcommand{\sig}{\sigma}

\title{The Calibration Channel Determines the Bayes-Error Proxy:\\
An Exact Law for Temperature-Induced Distortion}

\author{%
  Shreyas Pradeepkumar Khandale \\
  School of Computing \\
  Binghamton University (SUNY) \\
  Binghamton, NY, USA \\
  \texttt{skhandale@binghamton.edu} \\
}

\begin{document}

\maketitle

\begin{abstract}
The soft-label Bayes-error estimator $\beta(z) = \E[\min(z, 1-z)]$ of Ishida et al.\
[1] estimates the irreducible error of a binary task directly from probability-valued
labels. Recent work by Ushio et al.\ [2] showed that this estimator is fragile when the
probabilities are not the true posterior: even \emph{perfectly calibrated} soft labels
can yield a substantially inaccurate estimate, and they propose isotonic calibration as a
consistent remedy. We complement that line of work by characterizing \emph{exactly how}
the most widely used post-hoc calibration map---temperature scaling---distorts the proxy.
We prove an exact, model-free identity reducing the temperature-scaled proxy to the
classifier's margin distribution, from which we obtain (i) strict monotonicity in the
temperature and (ii) a continuous bijection from the temperature axis onto the open
interval $(0, \tfrac12)$, so that a \emph{fixed} classifier---with fixed decisions and
fixed $0$--$1$ error---can be made to report any proxy value whatsoever. Under a Gaussian
model of the logits we further derive a two-parameter closed form for the entire
proxy-versus-temperature curve. Across CIFAR-10, Fashion-MNIST, and SVHN (eight binary
tasks), the proxy varies by $56\times$ to $980\times$ at constant test error, the
closed form reproduces the empirical curve to within $0.018$, and the calibration
temperature that minimizes the expected calibration error does not coincide with any
stable proxy value. Our results give a precise, predictive account of the distortion
whose existence motivates calibration-based remedies, and they reinforce the practical
recommendation that a proxy value is meaningful only together with the mechanism that
produced its probabilities.
\end{abstract}

\section{Introduction}
\label{sec:intro}

When we evaluate a classifier, a basic question is whether its remaining errors can be
reduced at all, or whether they are intrinsic to the task. The Bayes error answers this.
It is the smallest error rate that any classifier can attain on a given problem, fixed
entirely by the data distribution rather than by the model one happens to choose. Two
uses follow. First, it serves as a yardstick: once a model's test error sits close to the
Bayes error, further architectural search buys little, because the remaining error is
irreducible. Second, it acts as a check on test-set overfitting---a reported test error
that falls below the Bayes error cannot come from an honest classifier and signals that
the test set has effectively leaked into training. The difficulty is that the Bayes error
is governed by the unknown class-conditional distributions, which makes it hard to
estimate in general, though by no means out of reach.

Building on the classical identity that the Bayes error of a binary task equals the
expectation of $\min\{\eta(x), 1-\eta(x)\}$ [7], where $\eta(x) = \Prob(y=1 \mid x)$ is
the true posterior probability of the positive class, Ishida et al.\ [1] turned it into
a practical estimator: replacing the expectation by a sample average, one averages
$\min\{c_i, 1-c_i\}$ over a collection of probability-valued \emph{soft labels} $c_i$.
Two features make the construction appealing. It is \emph{model-free}, in that no
classifier has to be trained, and \emph{instance-free}, in that only the probabilities
are needed and never the underlying inputs. The instance-free property is practically
useful---it allows estimation when the raw data are private or otherwise inaccessible,
and it avoids the dimensional scaling that burdens density-based estimators. We write the
functional at the heart of the estimator as
\begin{equation}
\beta(z) = \E[\min\{z, 1-z\}].
\label{eq:proxy}
\end{equation}
The estimator is unbiased precisely when the soft labels coincide with the true posterior;
that single conditional is the pivot on which everything in this paper turns.

In practice, though, the probabilities one feeds into \eqref{eq:proxy} are rarely the
oracle posterior. They arrive as human vote fractions, as the softmax outputs of a trained
network, or as the outputs of a post-hoc calibration map, and none of these is guaranteed
to equal $\eta$. Ishida et al.\ already saw a symptom of the gap: feeding the CIFAR-10H
human soft labels into the estimator produced a Bayes-error figure exceeding the test
error of a state-of-the-art classifier, which the true Bayes error can never do. Ushio et
al.\ [2] took up this distortion directly and established two things. First, calibrating
the soft labels is not by itself a remedy---soft labels that are calibrated to perfection
can still leave the estimate substantially inaccurate. Second, they showed that isotonic
calibration restores statistical consistency, provided the corruption preserves the
ordering of the underlying probabilities. Their analysis is what puts the existence of the
distortion, and a principled correction for it, on firm footing.

This paper complements [2] by sharpening the question and fixing attention on one
calibration map in particular. Temperature scaling [3]---dividing the logits by a positive
scalar $T$ before applying the sigmoid---is the default post-hoc calibration step for
modern neural networks. It carries one decisive property: dividing by $T$ is a monotone
transformation of the logits, so it can never move a point to the other side of the
decision boundary. The classifier's predictions, and with them its $0$--$1$ test error,
are left exactly as they were. This sets up a clean question. With the classifier frozen
in every respect that affects its accuracy, how does the proxy of \eqref{eq:proxy} respond
as $T$ sweeps across its range? We are after an exact, quantitative law---not merely the
fact that the proxy moves, but the precise functional form of how far and in which
direction.

\paragraph{Contributions.} We answer this exactly and predictively.
\begin{itemize}\itemsep1pt
\item \textbf{An exact, model-free identity} (Lemma~\ref{lem:fold}) reducing the
temperature-scaled proxy to the expectation of a simple function of the classifier's
\emph{margin} $|\ell(x)|$.
\item \textbf{A bijection theorem} (Theorem~\ref{thm:bij}, Corollary~\ref{cor:bij}): the
proxy is a continuous, strictly increasing bijection from the temperature axis onto
$(0, \tfrac12)$. A single fixed classifier---fixed decisions, fixed accuracy---can be made
to report \emph{any} proxy value in this range purely by choice of temperature.
\item \textbf{A Gaussian closed form} (Theorem~\ref{thm:closed}) for the entire
proxy-versus-temperature curve, depending on only two interpretable parameters (a
separation and a spread), with an explicit error bound (Proposition~\ref{prop:err}).
\item \textbf{Empirical validation} across CIFAR-10, Fashion-MNIST, and SVHN (eight binary
tasks): the proxy varies by $56\times$--$980\times$ at constant test error, and the closed
form matches the empirical curve to within $0.018$.
\end{itemize}
Our characterization gives the precise law for the distortion whose \emph{existence}
motivates the calibration-based estimators of [2], and it sharpens the practical message
that a Bayes-error proxy is interpretable only alongside the channel that produced its
probabilities.

\section{Setup}
\label{sec:setup}
Let $X$ be an input with binary label $Y \in \{0,1\}$, and let a fixed binary classifier
produce a logit $\ell(X) \in \R$. Post-hoc temperature scaling at temperature $T > 0$
produces the probability
\begin{equation}
p_T(x) = \sig\!\left(\frac{\ell(x)}{T}\right), \qquad \sig(u) = \frac{1}{1 + e^{-u}}.
\end{equation}
Since $\sig$ is strictly increasing and $\sig(0) = \tfrac12$, the decision
$h(x) = \mathbf{1}\{p_T(x) \ge \tfrac12\} = \mathbf{1}\{\ell(x) \ge 0\}$ is independent of
$T$; consequently the $0$--$1$ test error is invariant to $T$. We study the temperature
profile of the proxy \eqref{eq:proxy} evaluated on these probabilities,
\begin{equation}
\beta(T) = \E_X\!\big[\min\{p_T(X),\, 1 - p_T(X)\}\big].
\label{eq:betaT}
\end{equation}
We write $\Phi$ for the standard normal CDF and define the \emph{margin}
$M = |\ell(X)| \ge 0$.

\section{An exact law for temperature-induced distortion}
\label{sec:theory}
All proofs are deferred to Appendix~\ref{app:proofs}.

\begin{lemma}[Margin reduction]
\label{lem:fold}
For every $z \in \R$, $\min\{\sig(z), 1-\sig(z)\} = \sig(-|z|)$. Consequently the
temperature-scaled proxy is the exact, model-free expectation
\begin{equation}
\beta(T) = \E\!\left[\sig\!\left(-\frac{M}{T}\right)\right],
\label{eq:margin}
\end{equation}
where $M = |\ell(X)|$ is the margin.
\end{lemma}

Equation \eqref{eq:margin} is an identity: it involves no distributional assumption and
holds for any classifier. It is the reason the empirical proxy and the prediction
$\E[\sig(-M/T)]$ agree to numerical precision in our experiments (Section~\ref{sec:exp}).

\begin{theorem}[Monotone bijection]
\label{thm:bij}
If $\Prob(M > 0) > 0$ and $\Prob(M = 0) = 0$, then $\beta$ is continuous and strictly
increasing on $(0,\infty)$, with
\[
\lim_{T \to 0^+}\beta(T) = 0, \qquad \lim_{T \to \infty}\beta(T) = \tfrac12 .
\]
\end{theorem}

\begin{corollary}[A fixed classifier admits any proxy value]
\label{cor:bij}
Under the hypotheses of Theorem~\ref{thm:bij}, $T \mapsto \beta(T)$ is a continuous,
strictly increasing bijection from $(0, \infty)$ onto $(0, \tfrac12)$. Hence for any
target $\beta^\star \in (0, \tfrac12)$ there is a unique temperature $T^\star$ with
$\beta(T^\star) = \beta^\star$, while the classifier's decisions and $0$--$1$ test error
are unchanged for every $T$.
\end{corollary}

Corollary~\ref{cor:bij} is the formal statement that the calibration channel, not the
task alone, determines the proxy: holding the classifier fixed, the proxy can be steered
to any value in its range by temperature alone.

\begin{definition}[Gaussian logit model]
\label{def:gauss}
The logit follows a symmetric two-component mixture
$\ell \sim \tfrac12\,\mathcal N(\mu, s^2) + \tfrac12\,\mathcal N(-\mu, s^2)$, with
separation $\mu > 0$ and spread $s > 0$. We call $\mu/s$ the signal-to-noise ratio (SNR).
\end{definition}

\begin{theorem}[Gaussian closed form]
\label{thm:closed}
Under Definition~\ref{def:gauss} and the probit approximation
$\sig(u) \approx \Phi(\lambda u)$ with $\lambda = \sqrt{\pi/8}$, in the well-separated
regime $\mu \gg s$,
\begin{equation}
\beta(T) \;\approx\; \Phi\!\left(\frac{-\lambda \mu}{\sqrt{T^2 + \lambda^2 s^2}}\right).
\label{eq:closed}
\end{equation}
\end{theorem}

\begin{proposition}[Error bound]
\label{prop:err}
The error from the margin approximation used in Theorem~\ref{thm:closed} is at most the
per-class misclassification probability $\Phi(-\mu/s)$.
\end{proposition}

Equation \eqref{eq:closed} reproduces the high-temperature limit of
Theorem~\ref{thm:bij} ($\Phi(0) = \tfrac12$) and is accurate for $T \gtrsim s$. As
$T \to 0$, however, the closed form does not vanish but instead saturates at
$\Phi(-\mu/s)$, the model's per-class misclassification rate, whereas the exact proxy
$\beta(T) \to 0$. This is the regime in which the margin approximation of
Proposition~\ref{prop:err} is tight: the neglected left-tail contribution attains its
bound $\Phi(-\mu/s)$. In the well-separated regime this bound is numerically negligible,
so the absolute error stays small; but because the exact proxy itself vanishes, the closed
form becomes unreliable near $T=0$ in \emph{relative} terms, and the exact identity
\eqref{eq:margin} should be used there.

\section{Experiments}
\label{sec:exp}
\paragraph{Setup.} We train classifiers on CIFAR-10 [10], Fashion-MNIST [11], and
SVHN [12] and form balanced binary tasks following the construction of [1] (e.g.\ animals vs.\ artifacts on
CIFAR-10; upper-body garments vs.\ rest on Fashion-MNIST). For CIFAR-10 we additionally
train a five-member ResNet-18 [13] deep ensemble (test accuracy $96.0\%$), each member
temperature-scaled on a held-out split. For each binary task we sweep
$T \in [0.25, 5.0]$ and record the $0$--$1$ test error, the proxy $\beta(T)$ from
\eqref{eq:betaT}, the closed-form prediction \eqref{eq:closed} with $(\mu, s)$ fit to the
logits, and the expected calibration error (ECE). Implementation details are in
Appendix~\ref{app:details}.

\paragraph{The proxy varies by orders of magnitude at fixed accuracy.}
Table~\ref{tab:main} reports, per task, the (constant) test error and the range of the
proxy across the sweep. The test error is exactly invariant across all temperatures
(Section~\ref{sec:setup}), while the proxy spans factors of $56\times$ to $980\times$. The
proxy--error gap changes sign within every task: at small $T$ the proxy lies below the
test error (the model appears better than Bayes-optimal), and at large $T$ it lies far
above it.

\begin{table}[t]
  \caption{At fixed test error, the temperature-scaled proxy varies by up to $980\times$.
  ``Ratio'' is $\max_T \beta(T)/\min_T \beta(T)$, computed from unrounded endpoints. The
  test error is exactly constant across all temperatures within each row.}
  \label{tab:main}
  \centering
  \small
  \begin{tabular}{llccc}
    \toprule
    Dataset & Binary task & Test error (fixed) & $\beta(T)$ range & Ratio \\
    \midrule
    CIFAR-10      & animals vs.\ artifacts & 0.0145 & $[0.001458,\,0.1897]$ & $130\times$ \\
    CIFAR-10      & land vs.\ other        & 0.0173 & $[0.001873,\,0.1929]$ & $103\times$ \\
    CIFAR-10      & odd vs.\ even          & 0.0188 & $[0.002246,\,0.1954]$ & $87\times$ \\
    CIFAR-10      & first5 vs.\ last5      & 0.0339 & $[0.003638,\,0.2037]$ & $56\times$ \\
    Fashion-MNIST & garments vs.\ other    & 0.0015 & $[0.0001625,\,0.1593]$ & $980\times$ \\
    Fashion-MNIST & first5 vs.\ last5      & 0.0265 & $[0.001543,\,0.1789]$ & $116\times$ \\
    SVHN          & odd vs.\ even          & 0.0149 & $[0.001189,\,0.1843]$ & $155\times$ \\
    SVHN          & first5 vs.\ last5      & 0.0161 & $[0.001331,\,0.1850]$ & $139\times$ \\
    \bottomrule
  \end{tabular}
\end{table}

\begin{figure}[t]
  \centering
  \includegraphics[width=\linewidth]{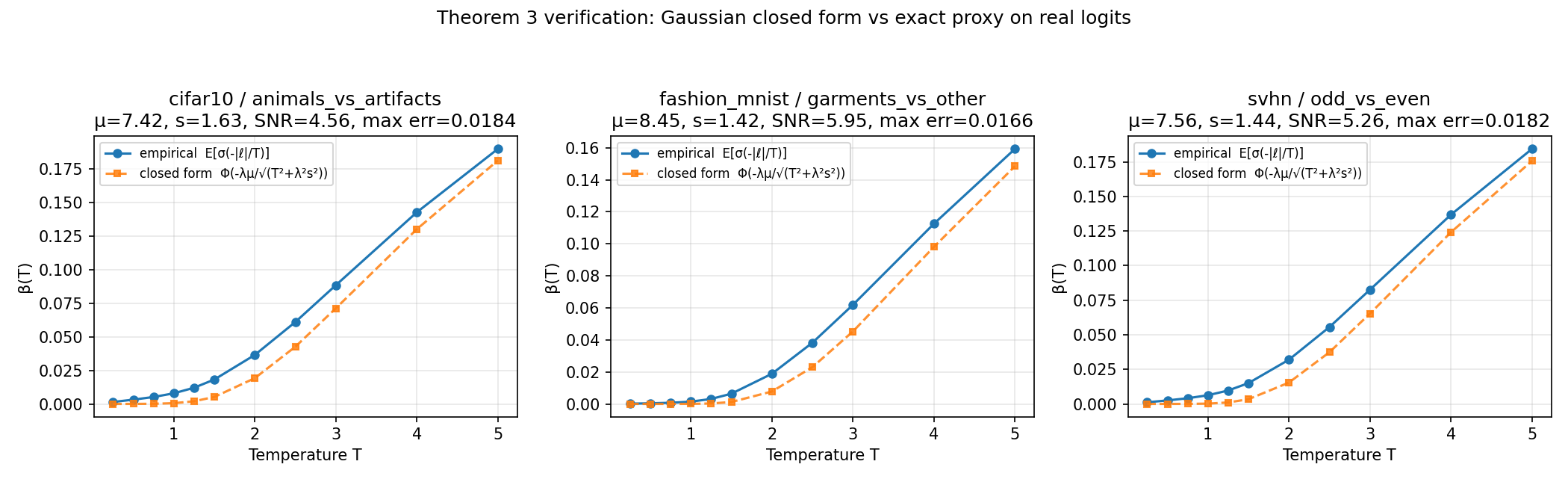}
  \caption{The two-parameter closed form \eqref{eq:closed} (dashed) tracks the exact
  proxy (solid) to within $0.018$ on all three datasets. The fitted SNRs lie in
  $\mu/s \in [4.6, 6.0]$, within the well-separated regime of Theorem~\ref{thm:closed};
  the residual matches the known maximum gap of the probit approximation.}
  \label{fig:closed}
\end{figure}

\paragraph{The closed form predicts the curve from two parameters.}
Fitting only $(\mu, s)$ to each task's logits, the closed form \eqref{eq:closed} tracks the
exact proxy with maximum absolute error $\le 0.018$ across all three datasets; the fitted
SNRs lie in $\mu/s \in [4.6, 6.0]$, within the well-separated regime of
Theorem~\ref{thm:closed}. The residual matches the known maximum gap of the probit
approximation ($\approx 0.018$) and shrinks at both temperature extremes
(Figure~\ref{fig:closed}).

\begin{figure}[t]
  \centering
  \includegraphics[width=\linewidth]{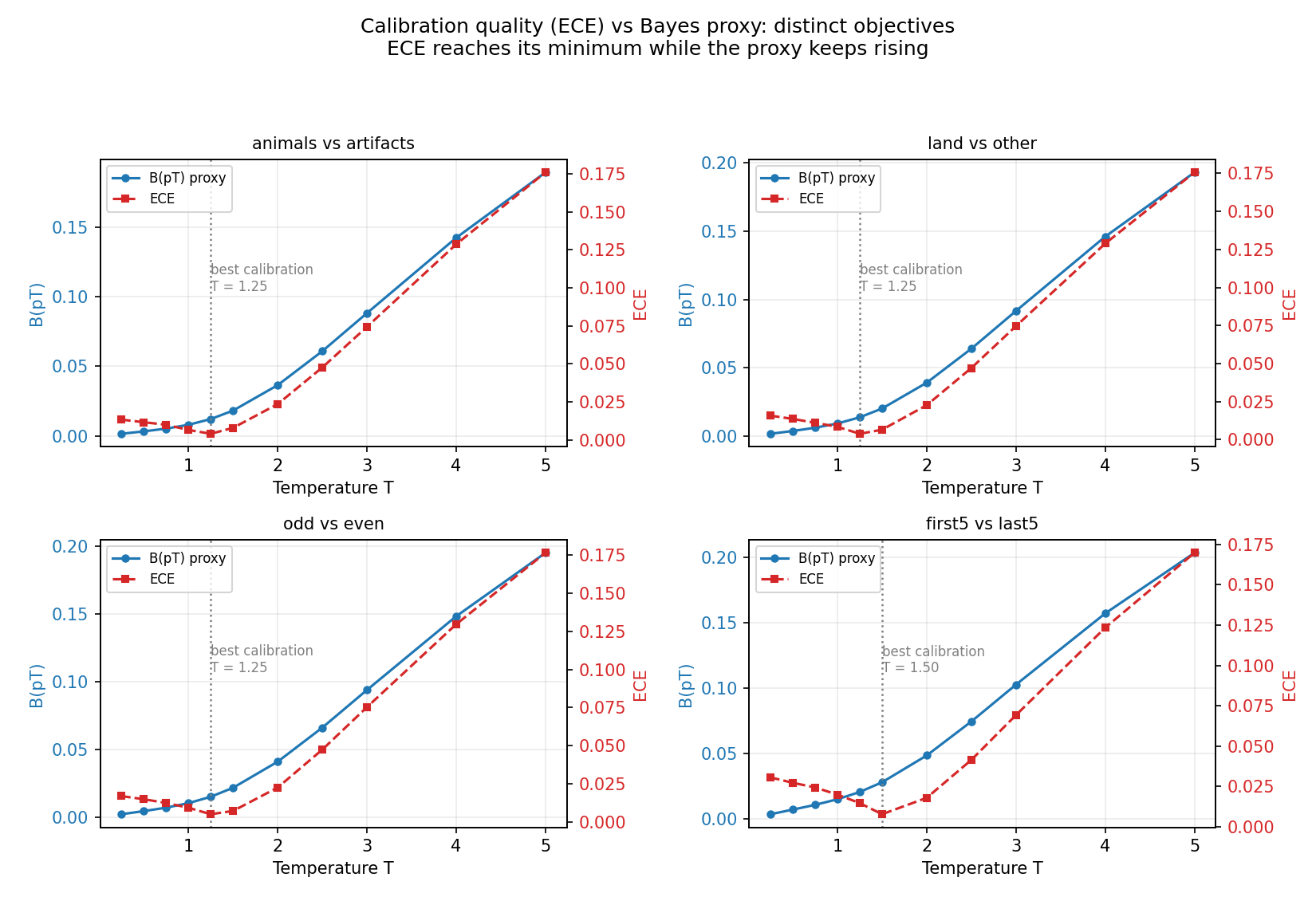}
  \caption{The proxy $\beta(T)$ keeps rising through the temperature that minimizes ECE;
  calibration quality and proxy stability do not coincide. Test error (not shown) is flat
  across all temperatures.}
  \label{fig:ece}
\end{figure}

\paragraph{Minimizing calibration error does not stabilize the proxy.}
The temperature that minimizes ECE (near $T \approx 1.25$ in our tasks) does not single out
a stable proxy value: the proxy rises monotonically straight through it
(Figure~\ref{fig:ece}). Calibration quality and proxy stability are distinct
objectives, consistent with the observation of [2] that calibration alone does not
guarantee an accurate estimate.

\paragraph{Channel dependence beyond temperature.}
The same effect appears across distinct probability sources, not only across temperatures.
At the NLL-optimal temperature, growing the CIFAR-10 ensemble from one to five members
raises the proxy on the animals-vs-artifacts task from $0.017$ to a plateau of $0.020$,
while the CIFAR-10H human vote-fraction channel gives $0.012$ on the same images: three
principled probability sources yield three distinct proxy values for one task. This
mirrors, from the model side, the human-soft-label distortion that motivates [2].

\section{Related work}
\label{sec:related}
Our work sits directly between two papers. Ishida et al.\ [1] introduced the soft-label
Bayes-error estimator \eqref{eq:proxy} and analyzed its bias under finite-annotator
averaging. Ushio et al.\ [2] showed that the estimator is unreliable under corrupted or
distribution-shifted soft labels---establishing that calibration alone is insufficient and
that even perfectly calibrated labels can mislead---and proposed isotonic calibration as a
consistent remedy, with an improved bias analysis for the hard-label estimator whose decay
rate adapts to class separation. Their treatment is concerned with \emph{whether} an
estimate is accurate and with constructing a consistent estimator. Ours is complementary
and orthogonal in emphasis: we fix a specific, ubiquitous calibration map (temperature
scaling) and derive the \emph{exact analytical law} by which it moves the proxy, including
a closed form and a bijection result that, to our knowledge, are not stated in either work.
Both works exploit class separation, for different quantities---their bias decay rate
versus our proxy curve in the well-separated regime. Multi-class extensions of the
soft-label estimator were developed by Jeong et al.\ [4]; instance-based Bayes-error
estimators form a separate, older line based on divergence estimation (e.g.\ Berisha et
al.\ [8]; Sekeh et al.\ [9]). Temperature scaling itself is due to Guo et al.\ [3], and
CIFAR-10H human soft labels to Peterson et al.\ [5]; deep ensembles follow
Lakshminarayanan et al.\ [6].

\section{Discussion and conclusion}
\label{sec:discuss}
Taken together, these results say that the temperature-scaled Bayes-error proxy obeys an
exact, model-free law. As the temperature varies it traces a strictly increasing bijection
onto the open interval $(0, \tfrac12)$; it admits a closed form in just two parameters, a
separation and a spread; and in our experiments it swings by as much as $980\times$---nearly
three orders of magnitude---on a classifier whose accuracy never moves. The bijection is
the conceptual core. Because every value in $(0, \tfrac12)$ is reachable while the
classifier stays fixed, the proxy cannot be read as a property of the task: holding the
model and its decisions constant, the choice of calibration temperature alone fixes what
the proxy reports. We note that the bijection itself is elementary---it follows from
pointwise monotonicity and the intermediate value theorem; its value is interpretive,
certifying that no proxy value is a task property, rather than technical. The decoupling
from calibration quality sharpens this. The temperature that minimizes expected calibration
error singles out no special proxy value---the proxy slides monotonically straight through
it---so one cannot recover a trustworthy estimate merely by calibrating well in the ECE
sense.

The contribution is deliberately diagnostic rather than constructive. We pin down the
distortion exactly, but we do not put forward a new estimator; for a corrected,
statistically consistent one we defer to Ushio et al.\ [2]. We read this as a clean
division of labor with their work rather than a gap in ours---their analysis supplies the
consistent remedy, ours supplies the exact law it corrects against, and the two are
complementary. The analysis is confined to binary classification; a multi-class treatment
is the natural next step, and the multi-class soft-label estimator of Jeong et al.\ [4] is
a sensible starting point, though it does not address the temperature characterization we
give here. We also treat a single calibration family. Temperature scaling is the most
common map, but isotonic, beta, histogram-binning, and Platt calibration may each distort
the proxy in their own way, and characterizing them remains open. Finally, our result has
two tiers: the margin identity of Lemma~\ref{lem:fold} is exact and model-free, whereas the
Gaussian closed form rests on a probit approximation that is tight only in the
well-separated regime.

The practical message is short. A Bayes-error proxy is only as meaningful as the process
that generated its probabilities; a bare figure such as $\beta = 0.05$ says nothing on its
own, because the same task can yield very different values depending on the annotator
protocol, the model, and the calibration map behind the number. We therefore recommend a
concrete change in practice: any reported soft-label Bayes-error estimate should come with
an explicit description of the probability-producing channel. By giving the exact law for
how one ubiquitous channel---temperature scaling---moves the proxy, this paper complements
the consistency-focused remedies of [2], and it turns a known hazard into something that
can be predicted and reported.

\section*{References}
\small
\begin{enumerate}\itemsep0pt
\item[{[1]}] T.~Ishida, I.~Yamane, N.~Charoenphakdee, G.~Niu, M.~Sugiyama.
  Is the performance of my deep network too good to be true? A direct approach to
  estimating the Bayes error in binary classification. \emph{ICLR}, 2023.
\item[{[2]}] R.~Ushio, T.~Ishida, M.~Sugiyama. Practical estimation of the optimal
  classification error with soft labels and calibration. \emph{arXiv:2505.20761}, 2025.
\item[{[3]}] C.~Guo, G.~Pleiss, Y.~Sun, K.~Q.~Weinberger. On calibration of modern neural
  networks. \emph{ICML}, 2017.
\item[{[4]}] M.~Jeong, M.~Cardone, A.~Dytso. Demystifying the optimal performance of
  multi-class classification. \emph{NeurIPS}, 2023.
\item[{[5]}] J.~C.~Peterson, R.~M.~Battleday, T.~L.~Griffiths, O.~Russakovsky. Human
  uncertainty makes classification more robust. \emph{ICCV}, 2019.
\item[{[6]}] B.~Lakshminarayanan, A.~Pritzel, C.~Blundell. Simple and scalable predictive
  uncertainty estimation using deep ensembles. \emph{NeurIPS}, 2017.
\item[{[7]}] T.~Cover, P.~Hart. Nearest neighbor pattern classification.
  \emph{IEEE Transactions on Information Theory}, 13(1):21--27, 1967.
\item[{[8]}] V.~Berisha, A.~Wisler, A.~O.~Hero, A.~Spanias. Empirically estimable
  classification bounds based on a nonparametric divergence measure. \emph{IEEE
  Transactions on Signal Processing}, 64(3):580--591, 2016.
\item[{[9]}] S.~Y.~Sekeh, B.~Oselio, A.~O.~Hero. Learning to bound the multi-class Bayes
  error. \emph{IEEE Transactions on Signal Processing}, 68:3793--3807, 2020.

\item[{[10]}] A.~Krizhevsky. Learning multiple layers of features from tiny images.
  Technical report, University of Toronto, 2009.
\item[{[11]}] H.~Xiao, K.~Rasul, R.~Vollgraf. Fashion-MNIST: a novel image dataset for
  benchmarking machine learning algorithms. \emph{arXiv:1708.07747}, 2017.
\item[{[12]}] Y.~Netzer, T.~Wang, A.~Coates, A.~Bissacco, B.~Wu, A.~Y.~Ng. Reading
  digits in natural images with unsupervised feature learning. \emph{NIPS Workshop on
  Deep Learning and Unsupervised Feature Learning}, 2011.
\item[{[13]}] K.~He, X.~Zhang, S.~Ren, J.~Sun. Deep residual learning for image
  recognition. \emph{CVPR}, 2016.
\end{enumerate}

\appendix

\section{Proofs}
\label{app:proofs}

\paragraph{Proof of Lemma~\ref{lem:fold}.}
Since $1 - \sig(z) = \tfrac{e^{-z}}{1+e^{-z}} = \tfrac{1}{1+e^{z}} = \sig(-z)$, we have
$\min\{\sig(z), 1-\sig(z)\} = \min\{\sig(z), \sig(-z)\}$. As $\sig$ is increasing, for
$z \ge 0$ the minimum is $\sig(-z) = \sig(-|z|)$, and for $z < 0$ it is
$\sig(z) = \sig(-|z|)$. Substituting $z = \ell(X)/T$ and $|\ell(X)/T| = M/T$ gives
\eqref{eq:margin}. \hfill$\square$

\paragraph{Proof of Theorem~\ref{thm:bij}.}
\emph{Monotonicity.} Fix $0 < T_1 < T_2$. For every $m > 0$, since $\sig$ is strictly
increasing and $-m/T_1 < -m/T_2$, we have $\sig(-m/T_1) < \sig(-m/T_2)$. Integrating this
strict pointwise inequality against the law of $M$ over $\{M > 0\}$, which carries positive
mass by hypothesis, gives $\beta(T_1) < \beta(T_2)$. No moment assumption on $M$ is needed.
\emph{Continuity and limits.} The integrand $\sig(-m/T)$ is bounded by $\sig(0) = \tfrac12$,
which is integrable against the law of $M$; continuity in $T$ then follows from dominated
convergence. As $T \to 0^+$, $\sig(-m/T) \to 0$ for every $m > 0$, so dominated convergence
and $\Prob(M=0)=0$ give $\beta(T) \to 0$; as $T \to \infty$, $\sig(-m/T) \to \tfrac12$ for
every $m$, giving $\beta(T) \to \tfrac12$. \hfill$\square$

\paragraph{Proof of Corollary~\ref{cor:bij}.}
Continuity follows from dominated convergence, strict monotonicity from
Theorem~\ref{thm:bij}, and the range $(0,\tfrac12)$ from the limits and the intermediate
value theorem. A continuous strictly increasing surjection onto an interval is a
bijection. \hfill$\square$

\paragraph{Proof of Theorem~\ref{thm:closed}.}
By symmetry of the mixture about $0$ and evenness of $\sig(-|\cdot|/T)$, the two components
contribute equally, so
$\beta(T) = \int_\R \sig(-|\ell|/T)\,\mathcal N(\ell;\mu,s^2)\,d\ell$. In the well-separated
regime the left-tail mass $\Phi(-\mu/s)$ is negligible, so $|\ell|$ may be replaced by
$\ell$ with error controlled by Proposition~\ref{prop:err}. Using the probit approximation
$\sig(-\ell/T) \approx \Phi(-\lambda \ell/T)$ and the Gaussian convolution identity
$\int \Phi(a + b\ell)\,\mathcal N(\ell;\mu,s^2)\,d\ell = \Phi\!\big((a+b\mu)/\sqrt{1+b^2 s^2}\big)$
with $a=0$, $b=-\lambda/T$, gives
$\beta(T) \approx \Phi\!\big(-\lambda\mu/\sqrt{T^2 + \lambda^2 s^2}\big)$ after multiplying
numerator and denominator by $T$. \hfill$\square$

\paragraph{Proof of the convolution identity.}
Let $Z \sim \mathcal N(0,1)$ be independent of $\ell \sim \mathcal N(\mu, s^2)$. Then
$\Phi(a + b\ell) = \Prob(Z \le a + b\ell \mid \ell)$, so the integral equals
$\Prob(Z - b\ell \le a)$. The variable $Z - b\ell$ is Gaussian with mean $-b\mu$ and
variance $1 + b^2 s^2$, giving $\Phi\!\big((a + b\mu)/\sqrt{1 + b^2 s^2}\big)$.
\hfill$\square$

\paragraph{Proof of Proposition~\ref{prop:err}.}
The integrands $\sig(-|\ell|/T)$ and $\sig(-\ell/T)$ agree on $\{\ell \ge 0\}$ and lie in
$[0,1]$ on $\{\ell < 0\}$, so their expected difference is at most
$\Prob(\ell < 0) = \Phi(-\mu/s)$ for the positive component. \hfill$\square$

\section{Experimental details}
\label{app:details}
All classifiers are ResNet-18 [13] adapted for $32\times32$ inputs (initial $3\times3$ stride-1
convolution, max-pool removed). CIFAR-10 ensemble members are trained for 200 epochs with
SGD (momentum $0.9$, weight decay $5\times10^{-4}$), cosine-annealed learning rate from
$0.1$, batch size $256$; per-member validation accuracy is $94.7$--$95.2\%$ and the ensemble
test accuracy is $96.0\%$. Each member is temperature-scaled on a held-out $5{,}000$-image
split by minimizing validation NLL (fitted $T^\star \approx 1.33$--$1.39$); the
ECE-minimizing temperature ($T \approx 1.25$) differs slightly, as NLL-optimal and
ECE-optimal temperatures need not coincide. Fashion-MNIST and SVHN use a single ResNet-18
trained $100$ epochs under the same optimizer. Binary logits are formed from the $10$-class
logits by the log-sum-exp difference of the positive and negative class groups. The
temperature grid is
$T \in \{0.25, 0.5, 0.75, 1.0, 1.25, 1.5, 2.0, 2.5, 3.0, 4.0, 5.0\}$. The proxy and its
analytical prediction \eqref{eq:margin} agree to within $10^{-9}$ across all $88$
task-temperature combinations. Code is available at
\url{https://github.com/sherurox/bayes-proxy-channel}.

\end{document}